# Emotion Detection using Image Processing in Python

Raghav Puri
Electronics & Communication Engineering
Bharati Vidyapeeth's College of Engineering
New Delhi, India
raghavpuri31@gmail.com

Archit Gupta
Electronics & Communication Engineering
Bharati Vidyapeeth's College of Engineering
New Delhi, India
archit.gupta33@gmail.com

Manas Sikri
Electronics & Communication Engineering
Bharati Vidyapeeth's College of Engineering
New Delhi, India
manassikri04@gmail.com

Mohit Tiwari,
Assistant Professor,
Bharati Vidyapeeth's College of Engineering
New Delhi, India
mohit.tiwari@bharatividyapeeth.edu

Nitish Pathak,
Assistant Professor,
BVICAM
New Delhi, India
nitishforyou@gmail.com

ShivendraGoel,
Associate Professor,
BVICAM
New Delhi, India
shivendragoel@gmail.com

*Abstract*—In this work, user's emotion using its facial expressions will be detected. These expressions can be derived from the live feed via system's camera or any pre-existing image available in the memory. Emotions possessed by humans can be recognized and has a vast scope of study in the computer vision industry upon which several researches have already been done. The work has been implemented using Python (2.7), Open Source Computer Vision Library (OpenCV) and NumPy. The scanned image (testing dataset) is being compared to training dataset and thusemotion is predicted. The objective of this paper is to develop a system which can analyze the image and predict the expression of the person. The study proves that this procedure is workable and produces valid results.

## I. INTRODUCTION TO IMAGE PROCESSING

In order to get an enhanced image and to extract some useful information out of it, the method of Image Processing can be used. It is a very efficient way through which an image can be converted into its digital form subsequently performing various operations on it. This is a technique similar to signal processing, in which the input given is a 2D image, which is a collection of numbers ranging from 0 to 255 which denotes the corresponding pixel value. [8]

TABLE 1: THIS IS A 2D ARRAY DEPICTING THE PIXELS OF A SAMPLE IMAGE

| 134 | 21 | 107 |
|-----|----|----|
| 64  | 37 | 78  |
| 42  | 4  | 13  |

The method involves converting an image into a 2D Matrix.

It consists of three basic steps [2]:

1.) **Scanning the image**: a raw image is acquired which has to be processed. It can be expressed in form of pixels as stated above. The aim of this step is to extract information which is suitable for computing.

2.) **Processing and Enhancing it**: -the image is converted into digital form by using a digitizer which samples and quantizes the input signals. The rate of sampling should be high for good resolution and high quantization level for human perception of different shades using different using gray-scale

3.) The obtained result describes the property of the image and further classifies the image.

**Conversion of Color Image to Gray Scale**

There are basically two methods to convert a color image to a gray scale image [8]:

A.) **Average Method**

In Average method, the mean is taken of the three colors i.e. Red, Blue & Green present in a color image. Thus, we get

Grayscale= (R+G+B)/3;

But what happens sometimes is instead of the grayscale image we get the black image. This is because we in the converted image we get 33% each of Red, Blue & Green.

Therefore, to solve this problem we use the second method called Weighted Method or Luminosity Method.

B) **Weighted or Luminosity Method**

To solve the problem in Average Method, we use Luminosity method. In these method, we decrement the presence of Red

Color and increment the color of Green Color and the blue color has the percentage in between these two colors.

Thus, by theequation [8],

Grayscale=((0.3 * R) + (0.59 * G) + (0.11 * B)).

We use this because of the wavelength patterns of these colors. Blue has the least wavelength while Red has the maximum wavelength.

II. Review of Literature

1. Different Strengths & Weaknesses were noted in the emotion recognition system by Mr. Ashwini, Mr. Jacob and Dr. Jubilant [14] of St. Joseph's College of Engineering.

2. Mr. ArunaChakraborty, Mr. Amit Konar, Mr. Uday Kumar Chakraborty, and Mr. Amita Chatterjee in. Emotion Recognition from Facial Expressions and Its Control Using Fuzzy Logic [15] explained that this fuzzy approach that has accuracy of about 90%.

3. Strengths and weaknesses of facial expression classifiers and acoustic emotion classifiers were analyzed by Carlos Busso, Zhigang Deng, SerdarYildirim, MurtazaBulut, Chul Min Lee, Abe Kazemzadeh, Sungbok Lee, Ulrich Neumann and Shrikanth Narayanan. [1].

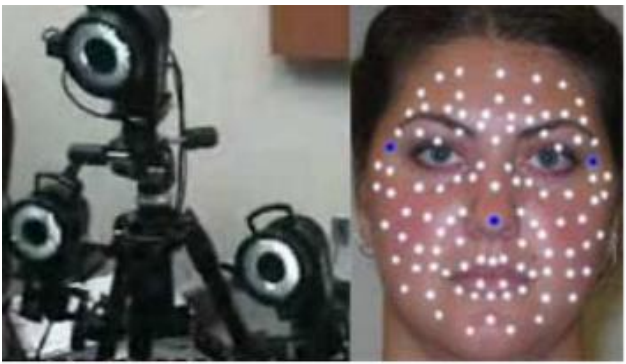

Fig.1 Data Recording System [16]

### III. Introduction to OpenCV

OpenCV is Open Computer Vision Library [4]. It is a free for all extensive library which consists of more than 2500 algorithms specifically designed to carry out Computer Vision and Machine Learning related projects. These algorithms can be put to use to carry out different tasks such as Face Recognition, Object Identification, Camera Movement Tracking, Scenery Recognition etc. It constitutes a large community with an estimate of 47,000 odd people who are active contributors of this library. Its usage extends to various companies both, Private and Public.

A new feature called GPU Acceleration [12] was added among the preexisting libraries. This new feature can handle most of the operations, though it's not completely advanced yet. The GPU is run by using CUDA and thus takes advantages from various libraries such as NPP i.e. NVIDIA performance primitives. Using GPU is beneficial by the fact that anyone can use the GPU feature without having a strong knowledge on GPU coding. In GPU Module, we cannot change the features of an image directly, rather we have to copy the original image followed by editing it.

Fig.2 Flowchart for Emotion Algorithm [13]

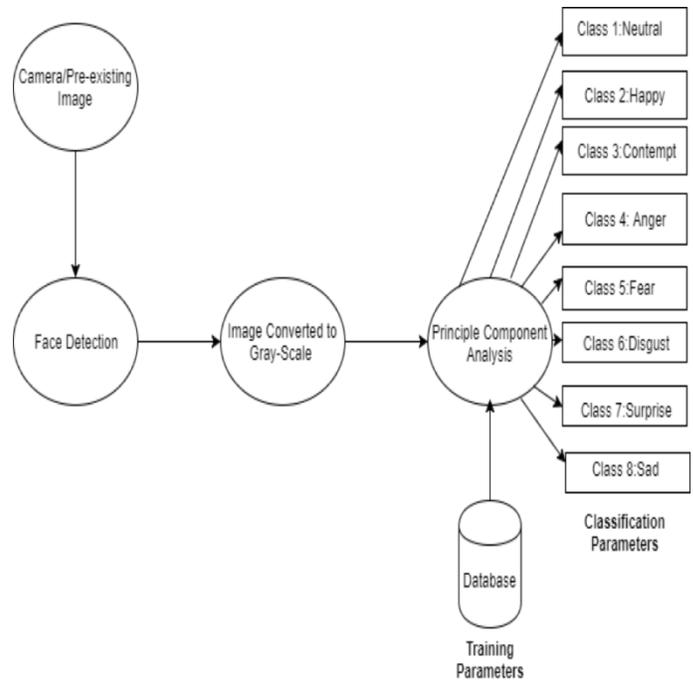

### I. IV. STEPS INVOLVED IN INSTALLING PYTHON 2.7 AND THE NECESSARY PACKAGES

Let's begin with a sample of image in either .jpg or .png format and apply the method of image processing to detect emotion out of the subject in the sample image. (The word 'Subject' refers to any living being out of which emotions can be extracted).

*A. Importing Libraries*

For successful implementation of this project, the following packages of Python 2.7 have to be downloaded and installed: Python 2.7.x, NumPy, Glob and Random. Python will be installed in the default location, C drive in this case. Open Python IDLE, import all the packages and start working.

*B. NumPy*

- NumPy is one of the libraries of Python which is used for complex technical evaluation. It is used for implementation of multidimensional arrays which consists of various mathematical formulas to process. [9].

- The array declared in a program has a dimension which is called as axis.

- The number of axis present in an array is known as rank

Emotion Detection using Image Processing in Python

- For e.g. A= [1,2,3,4,5]
- In the given array A 5 elements are present having rank 1, because of one-dimension property.
- Let's take another example for better understanding
- B= [[1,2,3,4], [5,6,7,8]]
- In this case the rank is 2 because it is a 2-dimensional array. First dimension has 2 elements and the second dimension has 4 elements. [10]

*C. Glob*

On the basis of the guidelines specified by Unix Shell, the Glob module perceives the pattern and with reference to it, generates a file. It generates full path name. [3]

1) Wildcards

These wildcards are used to perform various operations on files or a part of directory. There are various wildcards [5] which are functional out of which only two are useful: -

TABLE 2: Various files created in a Directory

| List of all files/working material saved inside a directory named "direc" |
|---|
| direc/filename1List of all files/working material saved inside a directory named "direc" |
| direc/filename1 |
| direc/filename2 |
| direc/filename3 |
| direc/filename4 |
| direc/filename5 |
| direc/files |

*a) Asterisk(*):* It represents any number of characters with any combination

For eg.
import glob
for name in glob.glob('direc/file*')
print name

Result=>
direc/filename1
direc/filename2
direc/filename3
direc/filename4
direc/filename5
direc/files

*b) Question Mark(?):* It represents or finds a single missing character

For e.g.
import glob

for name in glob.glob('direc/filename?')
print name

Result=>
    direc/filename1
direc/filename2
direc/filename3
direc/filename4
direc/filename5

This wildcard is limited to one specificdirectory and doesn't extend itself. i.e., itdoesn't find a file in a subdirectory.[5]

*D. Random*

Random Module picks or chooses a random number or an element from a given list of elements. This module supports those functions which provide access to such operations.

Classification of Random Module: -
1) randint(m,n)

    It returns a value of x such that-
    m<=x<=n

2) randrange(dog, cat, mouse, lion)

    It returns any random variable or element from the given range [7].

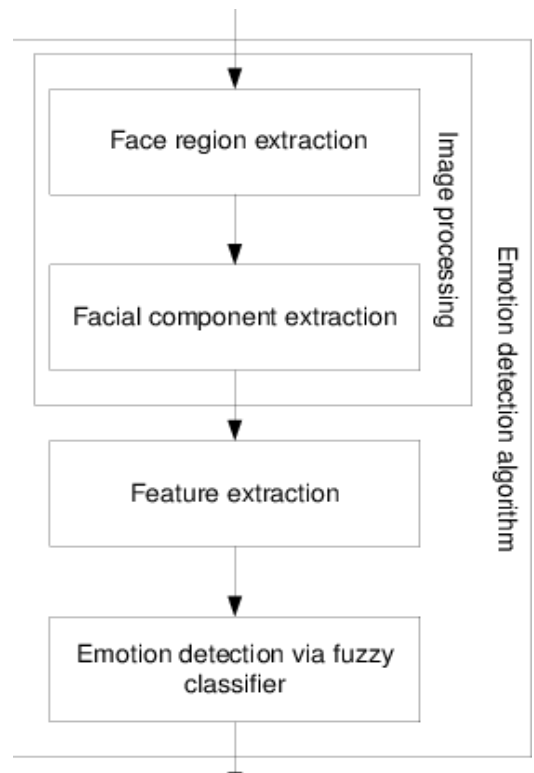

Fig.3 Flowchart for Emotion Detection

II. V. DIFFERENT EMOTIONS THAT CAN BE DETECTED OUT OF AN IMAGE:

A. A. Neutral

B. B. Happy

C. C. Anger

D. D. Disgust

E. Surprise

F. Fear

G. Sad

H. Contempt

III. VI. STEPS INVOLVED TO PERFORM EMOTION DETECTION USING OPENCV-PYTHON:

*1)* After successfully installing all the necessary softwares, we must start by creating a Dataset. Here, we can create our own dataset by analyzing group of images so that our result is accurate and there is enough data to extract sufficient information. Or we can use an existing database.

*2)* The dataset is then organized into two different
directories. First directory will contain all the images and the second directory will contain all the information about the different types of emotions.

*3)* After running the sample images through the
python code, all the output images will be stored into another directory, sorted in the order of emotions and its subsequent encoding.

*4)* Different types of classes can be used in OpenCV for emotion recognition, but we will be mainly using Fisher Face one. [1]

*5)* Extracting Faces:OpenCV provides four predefined classifiers, so to detect as many faces as possible, we use these classifiers in a sequence [1]

*6)* The dataset is split into Training set and Classification set. The training set is used to teach the type of emotions by extracting information from a number of images and the classification set is used to estimate the classifier performance.

*7)* For best results, the images should be of exact same properties i.e. size.

*8)* The subject on each image is analyzed, converted to grayscale, cropped and saved to a directory [1]

*9)* Finally, we compile training set using 80% of the test data and classify the remaining 20% on the classification set. Repeat the process to improve efficiency [1].

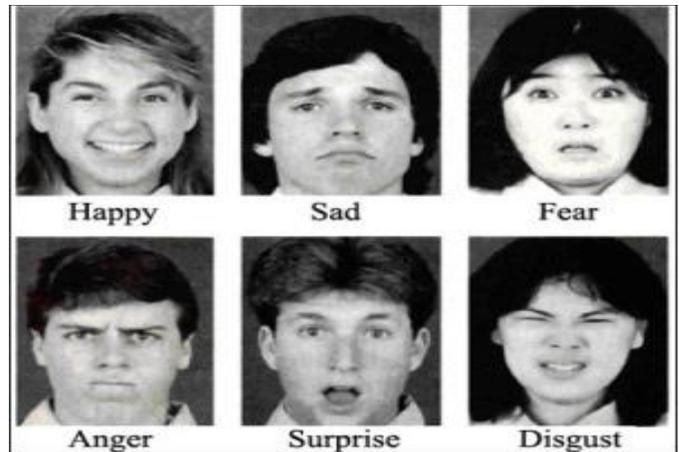

Fig.4 Output of a sample image [9]

VII. Cohn-Kanade AU-CODED EXPRESSION DATABASE

This is one of the Databases that can be used for emotion detection. It's a Database having numerous types of emotions [6]. This database is currently available in two different categories while the third one is under development stage.

The first version is labeled by "CK" while the second version is labeled as "CK+". The CK version starts by detecting the neutral emotion and then advances to the other higher emotions.

With the release of second type of this database, the frequency of processes increased by approximately 25% and the frequency of subjects by approximately 30% [6].

The third release is on the verge of development as which will have both the features of CK and CK+ Database, for e.g. adding integrated 30 degrees of rotation from the front.

IV. VIII. APPLICATIONS AND FUTURE SCOPE

Computer Vision is a very vast field which is still under developmental phase. Research work in this field is going at a rapid phase.

Emotion detection is an inseparable part of computer vision. Loads of tasks and processes can be performed if one can become aware about the intricacies and endless possibilities offered under the field of emotion detection.

Some common and widely used applications of emotion detection are:

A. App and product development

Emotion recognition can play a huge role in optimizing various software engineering processes which comprises of testing of ease with which a product can be used. It's a long-established fact that level of comfort with different software products depends hugely upon human emotions. A products overall look and feel can also alter human feelings which in turn makes the person buy the product or not. Thus, researching about different emotional states of a human body and how it is influenced by usage of different products is a matter of prime importance for related industry professionals.

### B. Improved learning practices

Evidence suggests that part of emotional states vouches for better learning practices while the other part try to suppress them. The difference between the two groups possessing different emotional states is not so common to find. For example, positive emotional state is thought to be bad for learning purposes while slightly negative emotional state fosters analytical thinking and is also appropriate for carrying out critical tasks.

### C. Improvised web development

With the mammoth scale at which the internet is expanding, service providers are interested in collecting tons and tons of data which can be extracted from the users. Correspondingly, all the content and advertisements are played based on the users' profile. Subsequently, adding intricate details about the different human emotions can provide much more precise behavioral models of different types of users.

### D. Immersive gaming

Video games constitute a large chunk of entertainment industry. Thus, in order to make these games much more intensive and intrusive, video game creators base their research on different types of human emotions commonly found. In order to allure more and more players, video games are made in such a way that they incorporate human emotions naturally into their game play.

IX. Challenges

The main purpose is to detect various emotions in a given sample image. The most challenging part in this task was to determine the exact emotion when two emotions look quite similar, for e.g. "Disgust" being classified as "Sadness", "Surprise" like "Happy" and so on. Now for eight different categories, the result was approximately 70% accurate which is quite well actually as our classifier learned quite a bit. So, we must see how can we increase its efficiency and accuracy.

If we look at our emotions list we can find out that we have only limited number of examples for "sad", "fear" and "contempt". By increasing the number of images for these emotions we can certainly increase optimization, or if we no longer consider these emotions in the list then optimization can be increased more than 80%.

The dataset that we used for our task was Cohn-Kanade (CK and CK+) Database. [6]

### V. X. CONCLUSION

Artificial Intelligence can be used to solve intriguing tasks such as emotion detection, although this task was quite convolute even more when using a great number of images. We humans also sometimes make a mistake while recognizing someone's emotion so is our program. The optimum accuracy was nearly 83%.

In Present Work, we recognized different types of human emotions using Python 2.7, OpenCV & (CK and CK+) Database [6] and got some interesting insight about it.